\documentclass[10pt,twocolumn,letterpaper,pagebackref,breaklinks,colorlinks,allcolors=iccvblue,hyphens]{article}

\usepackage{iccv}              

\usepackage{savesym}
\usepackage[subtle]{savetrees} 

\usepackage{graphicx}
\usepackage{subcaption}
\usepackage[font=small,labelfont=bf]{caption}
\usepackage{mwe}
\usepackage{amsmath}
\usepackage{tablefootnote}
\usepackage{listings}
\usepackage{array}
\usepackage{multirow}
\usepackage{hhline} 
\usepackage{tablefootnote}
\usepackage[accsupp]{axessibility}
\usepackage{bbding}
\usepackage{color}
 \usepackage[table]{xcolor}
\usepackage{cuted}
\usepackage{capt-of}
\usepackage{booktabs, multirow}
\usepackage{xcolor}
\usepackage{bbding}
\usepackage{svg}   
\usepackage{makecell}
\usepackage{pifont}

\usepackage{gensymb}
\usepackage{pgf}
\usepackage{kantlipsum}
\usepackage{cuted}
\usepackage{mathtools}
\newcommand{\rot}[1]{\rotatebox{90}{#1}}
\newcommand{\greencheck}{\textcolor{green!70!black}{\ding{51}}}
\newcommand{\redcross}{\textcolor{red}{\ding{55}}}
\usepackage{arydshln}
\usepackage{svg}  
\usepackage{pgf-pie}

\definecolor{iccvblue}{rgb}{0.21,0.49,0.74}

\def\paperID{12699} 
\def\confName{ICCV}
\def\confYear{2025}

\title{Towards Safer and Understandable Driver Intention Prediction}

\author{
Mukilan Karuppasamy\textsuperscript{1} \quad
Shankar Gangisetty\textsuperscript{1} \quad
Shyam Nandan Rai\textsuperscript{2} \quad
Carlo Masone\textsuperscript{2} \quad
C V Jawahar\textsuperscript{1} \\
\textsuperscript{1}\textit{IIIT Hyderabad, India}\quad
\textsuperscript{2}\textit{Politecnico di Torino, Italy}
}
\begin{document}
\maketitle
\begin{abstract}
Autonomous driving (AD) systems are becoming increasingly capable of handling complex tasks, mainly due to recent advances in deep learning and AI. As interactions between autonomous systems and humans increase, the interpretability of decision-making processes in driving systems becomes increasingly crucial for ensuring safe driving operations. Successful human-machine interaction requires understanding the underlying representations of the environment and the driving task, which remains a significant challenge in deep learning-based systems.
To address this, we introduce the task of interpretability in maneuver prediction before they occur for driver safety, i.e., driver intent prediction (DIP), which plays a critical role in AD systems. To foster research in interpretable DIP, we curate the eXplainable Driving Action Anticipation Dataset (\textbf{DAAD-X}), a new multimodal, ego-centric video dataset to provide hierarchical, high-level textual explanations as causal reasoning for the driver's decisions. These explanations are derived from both the driver's eye-gaze and the ego-vehicle's perspective. Next, we propose Video Concept Bottleneck Model (\textbf{VCBM}), a framework that generates spatio-temporally coherent explanations inherently, without relying on post-hoc techniques. Finally, through extensive evaluations of the proposed VCBM on the DAAD-X dataset, we demonstrate that transformer-based models exhibit greater interpretability than conventional CNN-based models. Additionally, we introduce a multilabel t-SNE visualization technique to illustrate the disentanglement and causal correlation among multiple explanations. Our data, code and models are available at: \href{https://mukil07.github.io/VCBM.github.io/}{\texttt{https://mukil07.github.io/VCBM.github.io/}}

\end{abstract}    
\section{Introduction}
\label{sec:intro}

\begin{figure}[t]
    \centering
    \includegraphics[width=\linewidth]{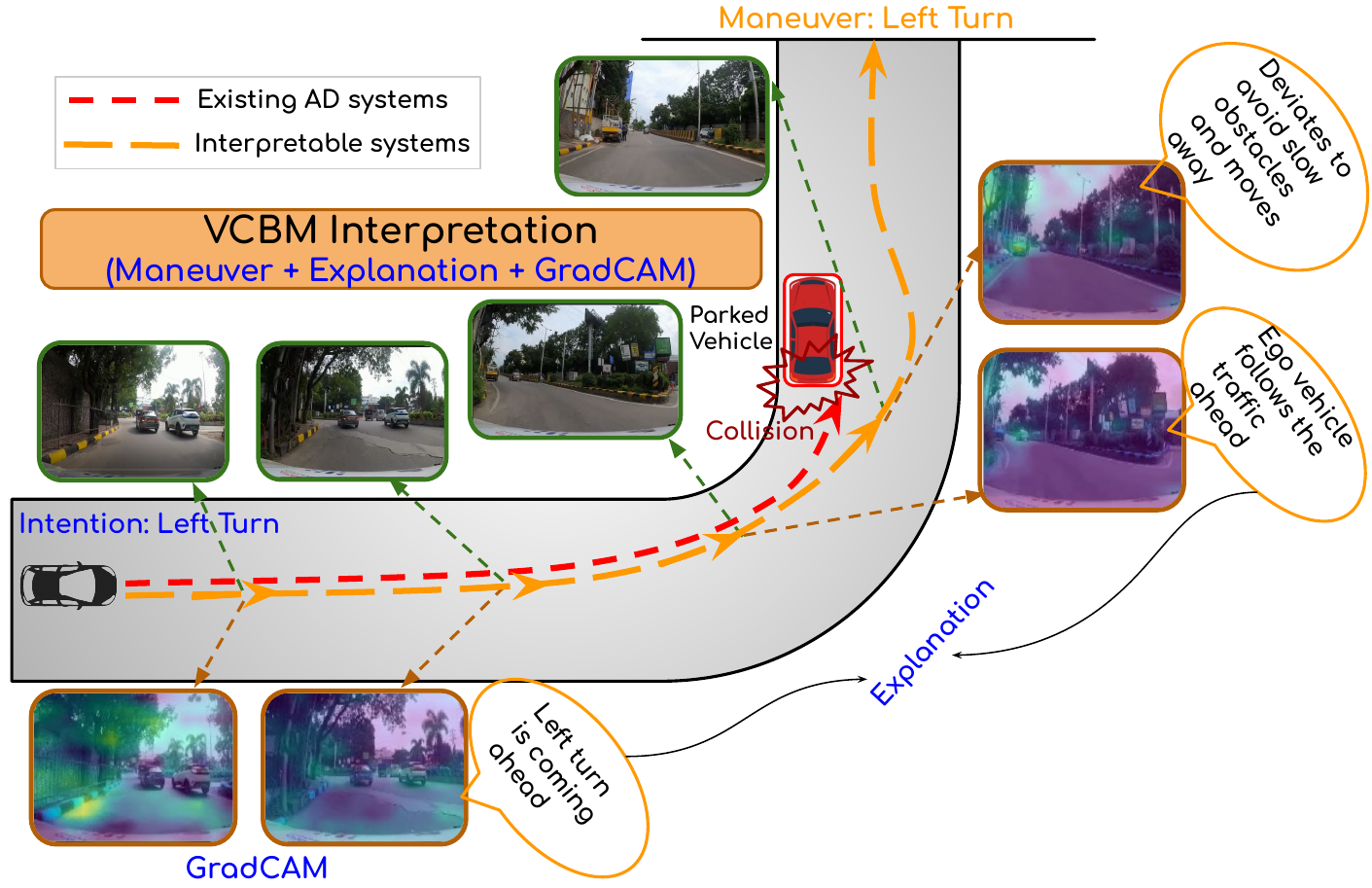}
    \caption{\textbf{Illustration of an AD scenario for the DIP task.} An AD system may intend to take a left turn while encountering a parked or slow-moving vehicle at the turn. Existing DIP models, lacking HCI understanding, might fail to anticipate the obstacle, leading to a potential collision. The proposed interpretable model, VCBM, enhances safety by enabling the ego-vehicle to explain its intended actions, anticipate obstacles more effectively, and adjust maneuvers accordingly. This results in safer and more transparent decision-making.}
    \label{fig:teaser_diagram}
\vspace{-.3cm}
\end{figure}

The increasing reliance on deep neural networks in safety-critical applications~\cite{Rudin} raises significant concerns due to their black-box nature, resulting in a lack of interpretability. In autonomous driving~\cite{li2019stereo,bddoia2020xu}, this lack of transparency makes it difficult for users to trust AI-driven decisions, leading to safety and accountability challenges, particularly in accidents. Ensuring safer deployment requires models that predict driving actions and provide human-understandable explanations for their decisions.

For instance, consider the scenario illustrated in Fig.~\ref{fig:teaser_diagram}. An autonomous car is traveling at high speed and attempts to take a left turn at a road intersection. While turning, a parked vehicle is in the blind spot and left undetected by the autonomous car sensors. In such situations, existing driver intention prediction (DIP) methods~\cite{daad24eccv,ma2023cemformer} may fail to recognize the parked vehicle, increasing the risk of a near-miss or collision. Hence, interpretability in DIP models becomes crucial in the aforementioned cases. Interpretable DIP models can reveal why the system overlooked such situations, which can help diagnose failures and improve model learning. 
An interpretable model can provide high-level explanations that enhance decision-making, fostering greater trust and confidence in autonomous driving technology. Trust is not solely about performance but also the ability to scrutinize, explain, and refine the model’s decisions over time, ultimately ensuring safer and more reliable deployment.

Traditional DIP datasets such as  Brain4Cars~\cite{jain2015car}, Viena{$^2$}~\cite{sadegh2018viena2}, HDD~\cite{ramanishka2018toward}, AIDE~\cite{yang2023aide}, and DAAD~\cite{daad24eccv} primarily focus on predicting maneuvers or agent trajectories without providing contextual explanations. This limitation affects the ability to train and evaluate DIP models on not just \textit{what} happened, but also \textit{why} it happened. To address this gap, we introduce the DAAD-X dataset (see Table~\ref{tab:dataset_comparison}), which includes both driving maneuvers (\textit{what}) and corresponding explanations (\textit{why}), enabling richer interpretability. 

However, due to their architectural limitations, existing DIP architectures cannot be directly employed to leverage such explanations effectively. For instance, recent architectures, such as VideoMAE~\cite{tong2022videomae}, DINOv$2$~\cite{Oquab2023DINOv2LR}, and MViTv$2$~\cite{li2022mvitv2} encode spatial and temporal information as flattened token representations, making it challenging to extract human-interpretable insights. Although self-supervised tasks such as frame ordering and motion prediction help capture temporal dynamics, the learned features often fail to correspond intuitively to human-understandable concepts. These limitations highlight the need for models that explicitly align learned features with explanations, ensuring both maneuver prediction and interpretability are jointly optimized.


We address this problem by including concept bottleneck models (CBM)~\cite{koh2020concept}, which are widely used to make models interpretable. CBM converts the highly uninterpretable features to low-dimensional, human-understandable explanations by training every neuron of a layer to represent one explanation. These explanations are fed to a sparse linear layer for the final model prediction. This results in a more straightforward interpretation of the final model prediction through linear combinations of interpretable explanations. However, applying these CBMs to video tasks is non-trivial since it does not understand the inherent temporal context of video data, a gap largely unexplored in the literature.


To overcome these challenges, we propose video CBM. By integrating spatially and temporally consistent tokens with CBMs, our approach delivers high-level explanations that naturally capture spatio-temporal features, offering the best of both worlds. To improve the understanding of DIP models, we make the following contributions in this work:
\begin{itemize}
    \item We propose DAAD-X, a multi-modal driving action anticipation video dataset incorporating hierarchical in-cabin eye-gaze and out-cabin ego-vehicle explanations. This dataset provides human-understandable justifications for driving maneuvers, enhancing interpretability and decision-making transparency. 
    \item Propose a multi-modal video-aware concept bottleneck model (VCBM) with learnable token merging and localized concept bottleneck. Our approach effectively leverages spatio-temporal features to disentangle explanations. To the best of our knowledge, this is the first work to propose a concept-based interpretability method explicitly tailored for video-based models.
    \item We presented qualitative results for VCBM on the DAAD-X dataset, demonstrating its improved performance across multiple backbone models. 
    In addition, we introduced a multi-label t-SNE visualization to highlight the causal correlation between multiple explanations in a video, offering a deeper interpretation of the model’s reasoning.
\end{itemize}

\begin{table}
\centering
\footnotesize
\setlength{\tabcolsep}{1.5pt}  
\def\arraystretch{1.2}
\newcolumntype{R}{>{\raggedleft\arraybackslash}p{1.4cm}}
\caption{\textbf{Comparison of datasets.} Our dataset is a subset of DAAD~\cite{daad24eccv} and it includes additional categories of explanations for multi-modal videos, encompassing both in-cabin (Aria eye-gaze) and out-cabin (ego-vehicle) perspectives.}
\begin{tabular}{R *{13}{c}}
\toprule
\multirow{1}{*}{{\textbf{Dataset}}} & \textbf{\rot{\makecell{\#In-cabin view}}} & \textbf{\rot{\makecell{\#Out-cabin view}}} & \textbf{\rot{Multimodal data}} &  \textbf{\rot{\makecell{Video data}}} & \textbf{\rot{Eye-gaze}} & \textbf{\rot{\makecell{eX-Annotation}}} & \textbf{\rot{\makecell{eX-Temporal\\   /Frame}}} & \textbf{\rot{eX-Scene}}  &
\textbf{\rot{\makecell{eX-PoV}}} & \textbf{\rot{\makecell{eX-Semantic}}} & \textbf{\rot{\makecell{eX-Causality}} } &

\\
\midrule

HDD~\cite{ramanishka2018toward} & 0 & 1 & No & Yes &  \redcross &  N/A &  N/A & \redcross & \redcross & \redcross & \redcross  \\ 

ROAD~\cite{road2021singh} & 0 & 1 & No & Yes & \redcross &  N/A &   N/A & \redcross & \redcross & \redcross & \redcross  \\ 

Dr(eye)~\cite{dreyeve2018palazzi} & 1 & 1 & Yes & Yes & \greencheck  &  N/A &   N/A& \redcross & \redcross & \redcross & \redcross  \\ 

DAAD~\cite{daad24eccv} & 2 & 4 & Yes & Yes &  \greencheck &  N/A & N/A & \redcross & \redcross & \redcross & \redcross \\ 

\hdashline[1pt/1pt]

BDD-OIA~\cite{bddoia2020xu} & 0 & 1 & No & No & \redcross & Categorical & Frame & \greencheck & \redcross & \redcross & \redcross  \\ 

BDD-X~\cite{kim2018bddx} & 0 & 1 & No & Yes &  \redcross & Contextual & Temporal & \greencheck & \redcross & \greencheck & \greencheck \\ 

\hdashline[1pt/1pt]
\textbf{DAAD-X (Ours)} & 2 & 4 & Yes & Yes &  \greencheck & Categorical & Temporal & \greencheck & \greencheck & \greencheck & \greencheck  \\ 

\bottomrule
\end{tabular}
\\ *eX means explanation

\label{tab:dataset_comparison}
\vspace{-.5cm}
\end{table}

\section{Related Work}
\label{sec:related_work}

\subsection{Driver Intention Prediction}
Various methods have been explored to recognize ego-vehicle actions and driver intentions. Early approaches, such as Hidden Markov Models~\cite{tran2015hidden}, focused on vehicle state prediction, while recent research has shifted to deep learning-based driver action anticipation. Traditionally, bidirectional RNNs~\cite{Olabiyi2017DriverAP} and CNN-LSTM architectures~\cite{Gebert2019EndtoendPO},~\cite{ jain2015car},~\cite{jain2016recurrent},~\cite{ khairdoost2020real}, ~\cite{rong2020driver},~\cite{Bonyani2023DIPNetDI} were used, though they often emphasize spatial features over temporal dependencies, limiting performance in extended video sequences. To address this, transformer-based architectures~\cite{vellenga2024evaluation} were introduced, improving long-range dependency capture, and memory-based anticipation methods such as Cemformer~\cite{ma2023cemformer} and M$^2$MVIT~\cite{daad24eccv} enhanced temporal consistency. One closely related work \cite{bddoia2020xu} produces explanations only for a single frame without incorporating the temporal context. The explanations are limited to short words or phrases, lacking the granularity required to capture cross-frame dynamics, making them unsuitable for interpreting video models.  

These video models remain uninterpretable, posing challenges for safe deployment in AD systems. To address this limitation, we propose a video-based interpretable intention prediction model with human-understandable explanations.


\subsection{Explainable Video Dataset}

Interpretability has recently gained significant attention, yet video-based interpretability remains challenging in tasks such as action recognition and long-video understanding~\cite{wang2024language,GarciaTorresFernandez2024InterpretabilityIV}. Existing DIP datasets, such as Brain4Cars~\cite{jain2015car}, Viena{$^2$}~\cite{sadegh2018viena2}, HDD~\cite{ramanishka2018toward}, AIDE~\cite{yang2023aide}, and DAAD~\cite{daad24eccv}, though they provide maneuver labels under diverse conditions, lack reasoning or explanatory annotations, limiting their suitability for interpretable models. BDD-OIA~\cite{bddoia2020xu} offers single frame-level explanations but fails to capture the spatio-temporal context necessary for comprehensive intention prediction models across a video. 
Although BDD-X \cite{kim2018bddx} offers detailed freeform contextual explanations, it cannot create interpretable models since we require categorical annotations to link a particular driving action to a specific, repeatable explanation with precise, distinct mapping. To bridge this gap, we introduce a new multi-modal video-based driving action dataset with human-understandable explanations to advance interpretability in autonomous driving.


\subsection{Concept-Based Explanations}
Understanding model decisions in multi-modal and temporal contexts is challenging due to the added temporal dimension and complex shared representations~\cite{ribeiro2016should}. Prior works, including concept bottleneck models~\cite{koh2020concept,oikarinen2023label}, and concept relevance propagation~\cite{achtibat2023attribution}, use fixed human-understandable concepts for decision-making, offering interpretability but failing to model temporal inputs~\cite{lee2024concept}. 
This limitation can lead to models learning spurious correlations and overlooking non-linear feature relationships. Due to the rigorous requirements of manual human-understandable annotations for CBM, the label-free CBMs were introduced in \cite{oikarinen2023label,tan2024explain,shang2024incremental} to generate concepts with the help of a pretrained text encoder. Recent methods such as LaIAR~\cite{wang2024language} and HENASY~\cite{vo2025henasy} have used language grounding on videos for contextual interpretability, but they fail in driving tasks due to the inability of the language model to capture positional and directional cues, which are essential in driving. To address this issue, we propose a simple framework that pools relevant features across frames, generating fine-grained, faithful explanations for videos.

\section{DAAD-X Dataset}
\label{sec:dataset}

\noindent\textbf{Motivation:} While driving on a straight, smooth road, a driver typically makes minimal steering adjustments or eye movements, as fewer decisions are required. However, during maneuvers—such as turning, lane changing, or stopping—the driver must be highly attentive, making precise hand movements based on visual cues. In such critical moments, DIP models can predict actions (e.g., turning left, slowing down), but they do not inherently explain why a particular action was predicted or whether it was the correct decision. For instance, consider a scenario where a driver approaches an intersection. If the DIP model predicts a left turn but does not indicate whether the decision was influenced by a traffic signal, movement of another vehicle , or presence of pedestrians, the prediction remains a black box. Without explanations, assessing whether the model's reasoning aligns with human decision-making is difficult.

To bridge this gap, it is essential to annotate DIP datasets with both actions and corresponding explanations, i.e., both ego-vehicle and eye-gaze explanations. By incorporating these explanations—such as the presence of obstacles, road signs, or the driver's eye-gaze behavior—we can develop interpretable models that not only predict actions and intentions but also justify their decisions. This would enhance trust, safety, and usability of autonomous driving.

\begin{figure}[h]
\centering
\includegraphics[width=\linewidth]{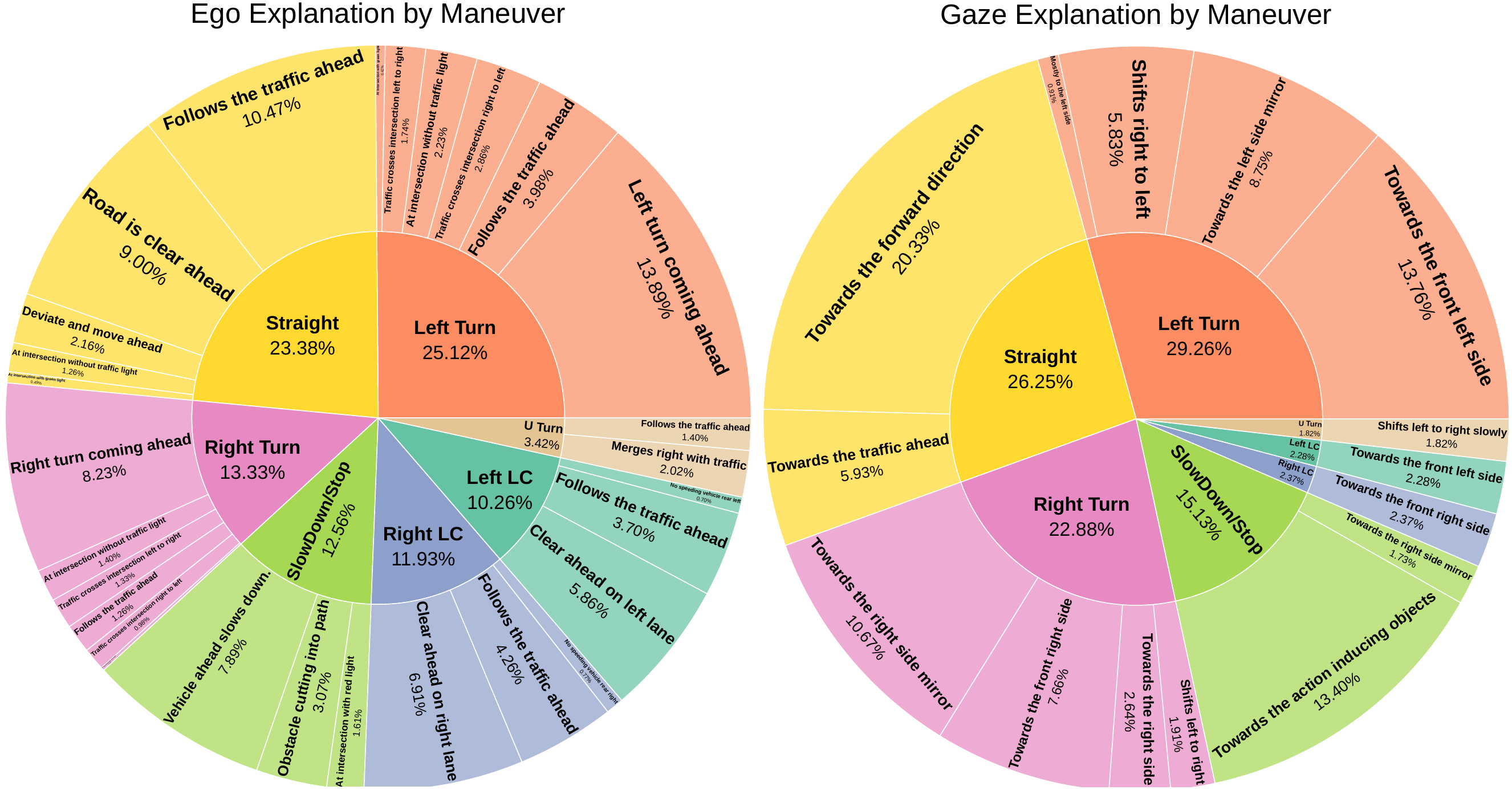}
    \caption{\textbf{Driving video annotation statistics of DAAD-X dataset.} Illustrating the distribution of (left) ego-vehicle explanations and (right) eye-gaze explanations across different maneuver actions. Details of the full explanation annotation are provided in the \textit{Supplementary Material}. Zoom in for better clarity.} 
    \label{fig:ego_explanations}
\end{figure}

\noindent\textbf{Dataset details:} To address these issues, we created a new dataset from the DAAD dataset~\cite{daad24eccv} to generate human-understandable explanations. DAAD was the closest match for our setup, as it is multi-modal with eye-gaze information and is well-conditioned across diverse weather conditions, drivers, times of day, and driving scenarios. The DAAD dataset includes seven intention labels, each corresponding to a specific maneuver: go straight (ST), right turn (RT), left turn (LT), right lane change (RLC), left lane change (LLC), slow/stop (SS), and U-turn (UT).
We selected $1,568$ video clips from the DAAD dataset, each ranging from $7$ to $15$ seconds in length. These videos are annotated with $17$ ego-vehicle explanations and $15$ gaze explanations using the VIA video annotator~\cite{Dutta19VIA}, an open-source tool. We refer to this enriched dataset as the explainable DAAD dataset (DAAD-X). Table~\ref{tab:dataset_comparison}
compares DAAD-X with previous datasets. 

\subsection{Data Annotation and Statistics}

\noindent\textbf{Annotation details:} 
During annotation, annotators watch each driving video and assign reasoning for the driver's maneuver. This process involves selecting a relevant gaze explanation and one or more ego-vehicle explanations to provide contextual justification. The gaze explanation is a single-attribute label chosen from $15$ predefined gaze explanations, which indicate where the driver is looking based on gaze coordinates collected using the Aria eye tracker~\cite{aria2023}. In contrast, ego-vehicle explanations consist of $17$ multi-attribute labels, where multiple explanations can be assigned to a single video. These explanations capture key scene attributes and offer semantically meaningful cues for both spatial and temporal localization. For example, in the explanations ``ego-vehicle is nearing the intersection", ``road is clear ahead on the left lane", ``gaze is mostly towards front right side", the term ``nearing" conveys temporal semantics, while ``clear ahead" and ``front right side" provide spatial context from the ego-vehicle's perspective.



\begin{figure*}[t]
    \centering
    \includegraphics[width=\linewidth]{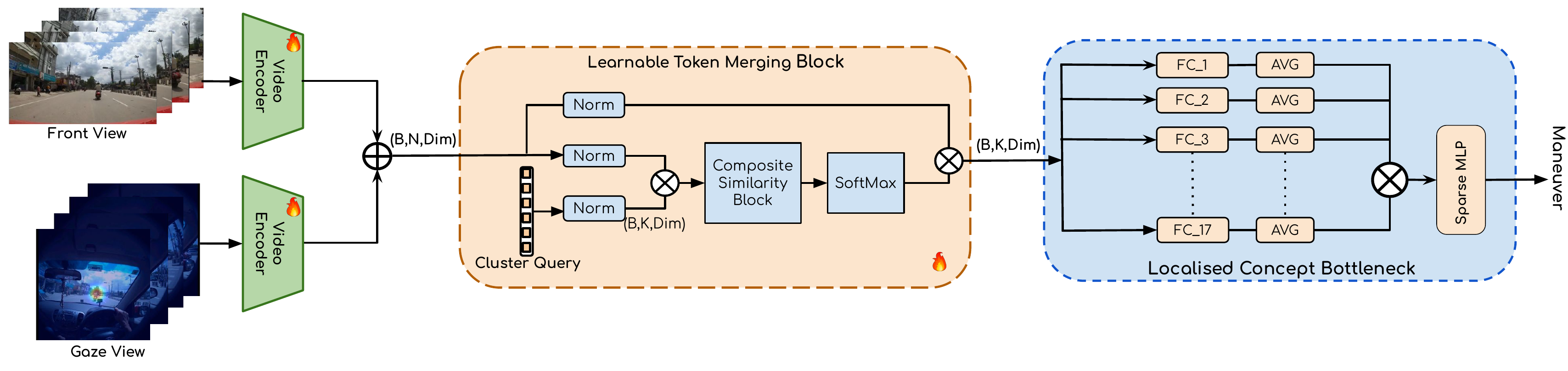}
    \caption{\textbf{Overall architecture of the proposed VCBM.} The dual video encoder first generates the spatio-temporal features (tubelet embeddings) for the ego-vehicle and gaze input sequence video pair. These tubelets are concatenated along the channel dimension and fed into the proposed learnable token merging block to produce $K$-cluster centers based on composite distances. These clusters are then fed into a localised concept bottleneck to disentangle and predict the maneuver label and one or more explanations to justify the maneuver decision.}
    \label{fig:proposed_architecture}
\end{figure*}

The annotated DAAD-X dataset provides a rich set of explanations, capturing both gaze and ego-vehicle attributes to enhance interpretability. As illustrated in Fig.~\ref{fig:ego_explanations}, each annotated instance includes the driver's maneuver, a gaze explanation, and multiple ego-vehicle explanations. 
In total, the dataset contains $2,536$ explanations, though their distribution is highly unbalanced. Among gaze explanations, the most frequent is ``towards the forward direction" ($223$ occurrences), while the least common is ``to the left side" ($10$ occurrences). Similarly, for ego-vehicle explanations, ``a left turn coming ahead" appears most frequently ($199$ occurrences), whereas explanations such as ``nearing an intersection and traffic light is green" are rare, occurring only $7$ times for the go straight maneuver, $6$ times for a left turn, and just $2$ time for a right turn. Given this long-tail distribution, we apply stratified sampling to ensure balanced representation, splitting the dataset into training ($70\%$), validation ($20\%$), and testing ($10\%$) sets.



\noindent
\textbf{Sanity Check.} Once annotated, the annotations were shuffled $3$ times among the annotators to validate the explanations. Since explanations are subjective, we initially selected the most obvious ones. For ambiguous cases, a consensus was reached based on the votes and comments of $10$ annotators. With this process, less than $1\%$ of the total videos were found to be incorrectly annotated and subsequently corrected. More details in \textit{Supplementary Material}.




\section{Video Concept Bottleneck Model (VCBM)}
\label{sec:method}







\subsection{Problem Formulation}
Given a set of input videos, where each video consists of \(x_g,x_f \in R^d \). \( x_g \) represents the gaze view video and \( x_f \) is a front view of an ego-vehicle. Now, we have a corresponding driving maneuvers prediction for each video sequence represented by \( y \) and explanations denoted as \( e \). \( e \in \{0,1\}^{17}\) represents the $17$ explanations. Consider the training dataset consisting of \( \{(x^i_g,x^i_f,y^i,e^i)\}_{i=1}^T \), where \(T\) is the total training instances. We can predict \( y = f(g(x))\) where \( g: R^d \rightarrow{R^{17}}\) represents the bottleneck layer, which maps input video features to $17$ intermediate explanations. $f : R^{17} \rightarrow{ R}$ is a sparse linear layer that maps intermediate explanations to the final maneuver prediction labels.

In our work, we introduce an unsupervised clustering module \(m : R^d \rightarrow{R^d}\) (detailed 
 discussion in Section \ref{sec:ltm}) to cluster similar features across frames. Finally, we follow \cite{koh2020concept} to learn bottleneck model \( (\hat{f},\hat{m},\hat{g})\) using the joint bottleneck approach, which minimizes the weighted sum as,  


\begin{equation}
\begin{split}
    \resizebox{0.5\textwidth}{!}{$\hat{f}, \hat{m}, \hat{g} = \arg\min_{f,m,g} \Biggl( 
\sum_i \left[ L_{Y}\bigl(f(g(m(x^{(i)}))), y^{(i)}\bigr) + \sum_j \lambda L_{C_j}\bigl(g(m(x^{(i)})), e_j^{(i)}\bigr)\right]
\Biggr)$}
\end{split}
\label{eq:loss}
\end{equation}

$L_y$  and $L_{C_j}$ represent multiclass cross entropy loss and multilabel aggregated binary cross entropy loss for each explanation $j \in \{1,17\}$. Here, $\lambda$ is the weighting factor.  




\subsection{Our Model Architecture}
We illustrate VCBM, our proposed model architecture, in Fig.~\ref{fig:proposed_architecture}. VCBM comprises a dual video encoder, a novel learnable token merging (LTM), and a localised concept bottleneck model (LCBM) module. LTM and LCBM help interpret both in-cabin gaze and out-cabin front video data effectively. At its core, VCBM predicts the driver's intended maneuver and provides human-understandable explanations for why the maneuver was selected, enhancing interpretability in the DIP task.


\noindent
\textbf{Video Encoder.} Our video encoder architecture is based on~\cite{vellenga2024evaluation}. For an input video sequence, gaze video as \( x_g^{i} \) and ego-vehicle front video as \( x_f^{i} \), we pass them individually to each branch to extract individual feature embeddings $z_i =(z_{g},z_{f}) \in R^{B\times N\times Dim}$. $B, N, Dim$ represents the batch size, number of tokens, and feature representation dimension. 
To retain spatial positioning while maintaining temporal consistency~\cite{zeng2024make}, we concatenate these feature embeddings along the channel dimension represented as $z'_i$. Now, we pass $z'_i$ through our proposed modules, Learnable Token Merging (LTM) and Localised Context Bottleneck Model (LCBM), to obtain the final prediction discussed in detail in the remaining section.


\subsection{Learnable Token Merging}\label{sec:ltm}
We introduce an LTM module to ensure the LCBM captures local features across frames. LTM groups semantically similar features into a reduced set of representative tokens, which are then passed as inputs to LCBM. We first perform unsupervised clustering in LTM on the concatenated multi-view feature representations $z^{'}_{i}$ from the encoder. The features are compared with \textit{$K$} learnable cluster centers $z_{c_j}$,  where $(i,j)$ represents token position and $K<<N$ to ensure that explanations are assigned to a compact set of merged features. The similarity between a feature token and a cluster center is computed using cosine similarity, given by,

\begin{equation}
    d_{feat}^{(i, j)} = 1 - \frac{z'_i \cdot z_{c_j}}{\lVert z'_i \rVert \lVert z_{c_j} \rVert}, \quad
\forall i \in \{1, \dots, N\}, \quad \forall j \in \{1, \dots, K\}
\end{equation}

We introduce a composite similarity block (in Fig.~\ref{fig:proposed_architecture}) that integrates and refines the similarity measures. The composite similarity block enhances clustering by enforcing spatial and temporal consistency. We compute additional spatial $\tilde{d}_{spatial}^{(i, j)}$ and temporal distances $ \tilde{d}_{temporal}^{(i, j)}$ ,


\begin{equation}
    \tilde{d}_{spatial}^{(i, j)} = \frac{d_{spatial}^{(i, j)}}{S_{\max}}, \quad 
    d_{spatial}^{(i, j)} = \sqrt{(x_i - x_{c_j})^2 + (y_i - y_{c_j})^2}
\end{equation}

\begin{equation}
 \tilde{d}_{temporal}^{(i, j)} = \frac{d_{temporal}^{(i, j)}}{T_{\max}}, \quad d_{temporal}^{(i, j)} = |t_i - t_{c_j}|
\end{equation}

The total composite distance used for clustering is, 

\begin{equation}
    d_{composite}^{(i, j)} = \alpha\, d_{feat}^{(i, j)} + \beta\, \tilde{d}_{spatial}^{(i, j)} + \gamma\, \tilde{d}_{temporal}^{(i, j)}
\end{equation}

Here, the $x_c,y_c,t_c$ represent the learnable cluster center position in spatial and temporal dimensions, respectively, and $\alpha, \beta, \gamma$ represent the normalized weights for distances. Instead of hard clustering \cite{Ester,Sinaga}, we employ soft clustering by assigning soft labels $w_{ij}$ to each token $z_i$ using a softmax over the negative composite distances,



\begin{equation}
    w_{ij} = \frac{exp(-d^{(i, j)}_{composite})}{\sum_{j=1}^{K}{exp(-d^{(i, j)}_{composite})}}
\end{equation}

The updated cluster centers are then computed as a weighted sum of token embeddings,

\begin{equation}
    \tilde{z}_{c_j} = \frac{\sum_{i=1}^{N}{w_{ij}z_i}}{\sum_{i=1}^{N}{w_{ij}}}
\end{equation}

By merging similar features into a compact token representation, this approach reduces redundancy in video embeddings while ensuring that spatio-temporally relevant features are retained. These merged token representations serve as inputs to the LCBM, enabling it to generate fine-grained explanations while maintaining spatial and temporal consistency.

\subsection{Localised Context Bottleneck Model}

\begin{figure}[h]
    \centering
    \includegraphics[width=\linewidth]{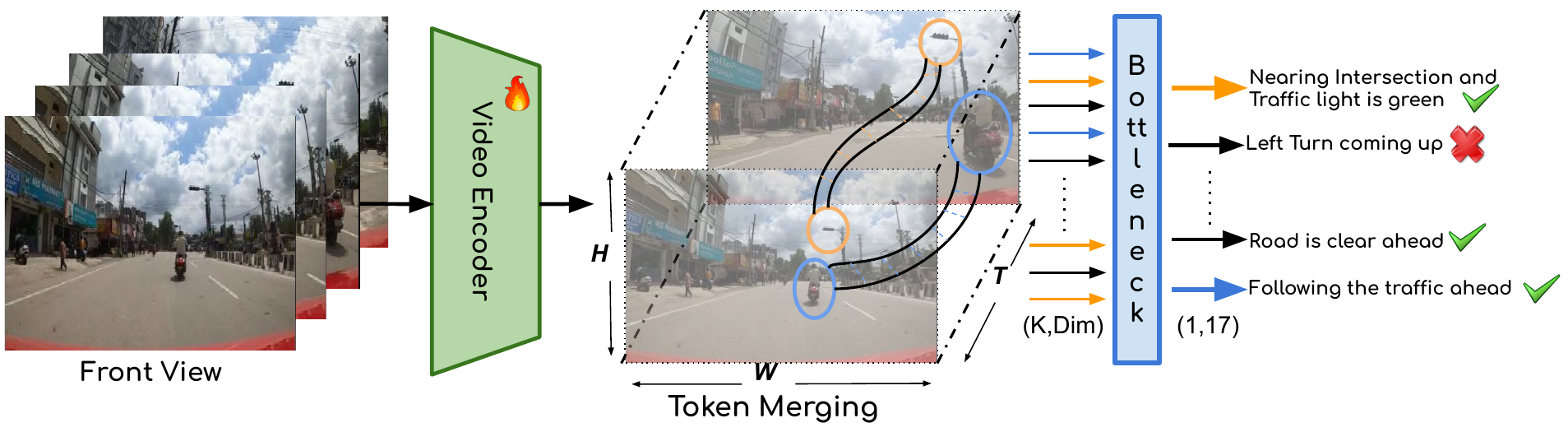}
    \caption{
    \textbf{VCBM merges relevant features across frames $(z_{c_j})$ and assigns explanations. } \textcolor{blue}{\textbf{Blue}} represents \textcolor{blue}{\textbf{merged traffic features}}, \textcolor{orange}{\textbf{orange}} denotes \textcolor{orange}{\textbf{merged traffic light features}}, and arrow thickness indicates prediction confidence.
    }
    
    \label{fig:lbcm}
\end{figure}

The LCBM further refines the representations from LTM by mapping high-dimensional encoded vectors to a human-understandable low-dimensional space. Traditional CBM approaches rely on global feature embeddings or global average pooling, which can discard fine-grained spatial and temporal details. Instead, LCBM approach preserves these details by feeding all the pooled token representations into the bottleneck block ($g(z_c)$), as illustrated in Fig.~\ref{fig:lbcm}.

Rather than immediately averaging features before the bottleneck, we introduce a late averaging strategy, allowing each merged token to retain its individual contribution to the explanation process. Each fully connected (FC) layer in the bottleneck module corresponds to a specific explanation and produces a single logit, representing the confidence of that explanation. This ensures that each FC layer processes all tokens, enabling a more robust and interpretable assignment of explanations. By retaining fine-grained spatio-temporal details, this LCBM enhances the activation maps, leading to more precise and human-understandable explanations in DIP.


\section{Experiments}
\label{sec:experiments}

\subsection{Implementation Details}
\noindent
 For our experiments, we used I3D~\cite{carreira2018quovadisactionrecognition} pre-trained on ImageNet RGB images, as well as VideoMAE~\cite{tong2022videomae} with a ViT-B/$16$ backbone~\cite{dosovitskiy2020image} and MViTv$2$-B~\cite{li2022mvitv2}, both pre-trained on Kinetics-$400$ dataset. For more details on data augmentation, training parameters, and evaluation metrics refer \textit{Supplementary Material}. 

\subsection{Results}

\begin{table}
\centering
\caption{\textbf{Evaluation on DAAD-X dataset:} Evaluated baselines with (wB) and without (woB) bottleneck. Here, LTM indicates Learnable Token Merging.}
\label{tab:model_performance_across_dataset}
\footnotesize 
\setlength{\tabcolsep}{2pt} 
\renewcommand{\arraystretch}{1.5}
\begin{tabular}{c |cc | cc | cc}
\hline
\multirow{2}{*}{\textbf{Model}} & \multicolumn{2}{c|}{\textbf{Action}} & \multicolumn{4}{c}{\textbf{ego-vehicle eXplanation}} \\
\cline{2-7}
& \multicolumn{1}{c}{Acc} & \multicolumn{1}{c|}{$F_1$} & \multicolumn{1}{c}{Acc} & \multicolumn{1}{c|}{$F_1$} & \multicolumn{1}{c}{$F_1(mac)$} & \multicolumn{1}{c}{$F_1(mic)$} \\
\hline
I3D woB ~\cite{carreira2018quovadisactionrecognition} & 74.78 & 74.21 & - & - & - & - \\
VideoMAE woB ~\cite{vellenga2024evaluation} & 72.5 & 71.81  & - & - & - & - \\
MViTv2 woB~\cite{li2022mvitv2} & 64.03 & 63.98  & - & - & - & - \\
\hdashline[1pt/1pt]
I3D wB~\cite{carreira2018quovadisactionrecognition} & 74.09 & 73.47 & 25.26 & 36.73 & 18.53 & 43.49 \\
VideoMAE wB~\cite{vellenga2024evaluation} & 67.01 & 66.48 & 24.21 & 38.24 & 23.77 & 41.53 \\
MViTv2 wB~\cite{li2022mvitv2} & 63.29 & 62.47 & 25.35 & 37.1 & 24.3 & 42.1 \\
\hdashline[1pt/1pt]
\textbf{I3D + LTM wB} \textbf{(Ours)} & \textbf{73.21} & \textbf{72.2} & \textbf{28.31} & \textbf{39.43} & \textbf{24.1} & \textbf{44.06} \\
\makecell{\textbf{MViTV2 + LTM wB}\\ \textbf{(Ours)}} & \textbf{69.73} & \textbf{69.15} & \textbf{31.22} & \textbf{43.86} & \textbf{29.17} & \textbf{49.11} \\
\hline
\end{tabular}
\end{table}

We compare our proposed method with three backbone architectures: the CNN-based I3D, transformer-based VideoMAE, and MViTv$2$. Table~\ref{tab:model_performance_across_dataset} shows the performance of the baselines and the backbone models with and without the bottleneck layer. We observe that the transformer-based MViTv$2$ baseline outperforms the CNN-based I3D baseline in predicting explanations using the bottleneck layer. While CNN excels in spatial feature extraction, video-based explanation tasks require a strong temporal understanding across frames, making transformers more effective.

\subsection{Insights and Ablations}
In this section, we provide additional insights into the effect of token merging, the importance of gaze modality in VCBM, and the impact of temporal cues.

\begin{table}
\setlength{\tabcolsep}{2pt} 
\renewcommand{\arraystretch}{1.5}

\caption{\textbf{Importance of token aggregation.} Compared the effectiveness of token merging (using all the tokens from the encoder) over utilizing the CLS token for producing explanations.}
\centering
\footnotesize 
\begin{tabular}{c | c |cc | cc | cc}
\hline
  \multirow{2}{*}{\makecell{\textbf{Strategy}}} & \multirow{2}{*}{\makecell{\textbf{Model}}} & \multicolumn{2}{c|}{\textbf{Action}} & \multicolumn{4}{c}{\textbf{ego-vehicle eXplanation}} \\
\cline{3-8}
& & \multicolumn{1}{c}{Acc} & \multicolumn{1}{c|}{$F_1$} & \multicolumn{1}{c}{Acc} & \multicolumn{1}{c|}{$F_1$} & \multicolumn{1}{c}{$F_1(mac)$} & \multicolumn{1}{c}{$F_1(mic)$} \\
\hline
  \multirow{2}{*}{\shortstack{CLS Token\\Summarization}} & I3D woB & 74.38 & 74.1 & - & - & - & - \\
  &   I3D wB & 73.73 & 72.25 & 23.15 & 33.7 & 16.9 & 41.95 \\
\hline
 \multirow{2}{*}{\shortstack{Full Token\\Aggregation}*} & I3D woB & \textbf{74.78} & \textbf{74.21} & -& - & - & -\\
   & I3D woB & 73.21 & 72.2 & \textbf{28.31} & \textbf{39.43} & \textbf{24.1 }& \textbf{44.06}\\ 
\hline
\end{tabular}
\label{tab:aggregation}
\end{table}

\subsubsection{Effects of LTM and LCBM}

\begin{table}
\centering
\caption{\textbf{Effect of number of clusters.} A bottleneck with lower clusters learns more global representation. Increasing the clusters further reduces performance due to noisy cluster centers.}
\label{tab:cluster}
\footnotesize 
\setlength{\tabcolsep}{2pt} 
\renewcommand{\arraystretch}{1.2}
\begin{tabular}{c | c |cc | cc | cc}
\hline
\multirow{2}{*}{\textbf{Model}}  & \multirow{2}{*}{\textbf{\#Clusters}}  & \multicolumn{2}{c|}{\textbf{Action}} & \multicolumn{4}{c}{\textbf{ego-vehicle eXplanation}} \\
\cline{3-8}
& & \multicolumn{1}{c}{Acc} & \multicolumn{1}{c|}{$F_1$} & \multicolumn{1}{c}{Acc} & \multicolumn{1}{c|}{$F_1$} & \multicolumn{1}{c}{\makecell{$F_1$\\$(mac)$}} & \multicolumn{1}{c}{\makecell{$F_1$\\$(mic)$}} \\
\hline
I3D + LTM wB  & 1 & 70.78 & 70.47 & 24.64 & 34.47 & 16.67 & 42.29 \\
I3D + LTM wB  & 3 & 71.78 & 71.44 & 25 & 38.46 & 22.85 & 43.87 \\
I3D + LTM wB & 5 & 73.21 & 72.2 & \textbf{28.31} & \textbf{39.43} & \textbf{24.1} & \textbf{44.06} \\
I3D + LTM wB & 7 & \textbf{74.28} & \textbf{74.03} & 24.64 & 34.9 & 18.44 & 41.01 \\
I3D + LTM wB  & 10 & 70.35 & 69.64 & 23.92 & 33.28 & 16.09 & 41.2 \\
\hdashline[1pt/1pt]
MViTV2 + LTM wB   & 5 & \textbf{69.73} & \textbf{69.15} & \textbf{31.22} & \textbf{43.86} & \textbf{29.17} & \textbf{49.11} \\
MViTV2 + LTM wB   & 10 & 65 & 64.53 & 30 & 43.51 & 27.11 & 47.11 \\
\hline
\end{tabular}
\end{table}

In Table~\ref{tab:cluster}, we analyze the impact of varying the number of clusters in the LTM block (see Fig.~\ref{fig:proposed_architecture}) on both action and explanation predictions in the proposed method. Using a single cluster is analogous to employing a \texttt{CLS} token in a transformer, where all tokens are aggregated into one global representation. As the number of clusters increases, each cluster is attributed to certain similar features across the tokens, but adding more clusters counteracts by learning additional noise patterns, which detracts from the prediction performance. 

LCBM is designed to compute explanations based on all locally aggregated tokens from the LTM block, rather than relying on a single global \texttt{CLS} token. This approach enhances the justification of explanations by preserving the contextual integrity of feature groups, making the model more interpretable as illustrated in the Table \ref{tab:aggregation}. The component-level ablation shown in Table ~\ref{tab:component_wise} demonstrates that incorporating the LCBM block improves explanation and action performance metrics. By attending to all the input tokens, the LCBM effectively retains fine-grained details, leading to more precise bottleneck representations. Additionally, integrating the LTM block further enhances explanation performance due to its ability to extract meaningful merged tokens. However, the averaging process during token merging reduces the granularity of individual features, resulting in a slight drop in action prediction accuracy. 

\begin{table}
\centering
\caption{\textbf{Component-level ablation.} Significance of proposed modules (LTM and LCBM) on I3D architecture.}
\label{tab:component_wise}
\footnotesize 
\setlength{\tabcolsep}{3pt} 
\renewcommand{\arraystretch}{1.2}
\begin{tabular}{ cc |cc | cc | cc}
\hline
 \multicolumn{2}{c|}{\textbf{Components}}  & \multicolumn{2}{c|}{\textbf{Action}} & \multicolumn{4}{c}{\textbf{ego-vehicle eXplanation}} \\
 \cline{1-8}
 \multicolumn{1}{c}{LTM} & \multicolumn{1}{c|}{LCBM} & \multicolumn{1}{c}{Acc} & \multicolumn{1}{c|}{$F_1$} & \multicolumn{1}{c}{Acc} & \multicolumn{1}{c|}{$F_1$} & \multicolumn{1}{c}{$F_1(mac)$} & \multicolumn{1}{c}{$F_1(mic)$} \\
\hline
\redcross & \redcross & 68.1 & 67.44 & 11.22 & 21.44 & 9.37 & 22.51 \\
 \greencheck & \redcross & 72.8 & 72.15 & 26.03 & 35.6 & 19.15 & 44.1 \\
\redcross & \greencheck & \textbf{74.09} &\textbf{ 73.47} & 25.26 & 36.73 & 18.53 & 43.49 \\
\greencheck & \greencheck &  73.21 & 72.2 & \textbf{28.31} & \textbf{39.43} & \textbf{24.1} & \textbf{44.06} \\
\hline
\end{tabular}
\end{table}

\subsubsection{Importance of Gaze modality}

\begin{figure}[h]
    \centering
    \includegraphics[width=\linewidth]{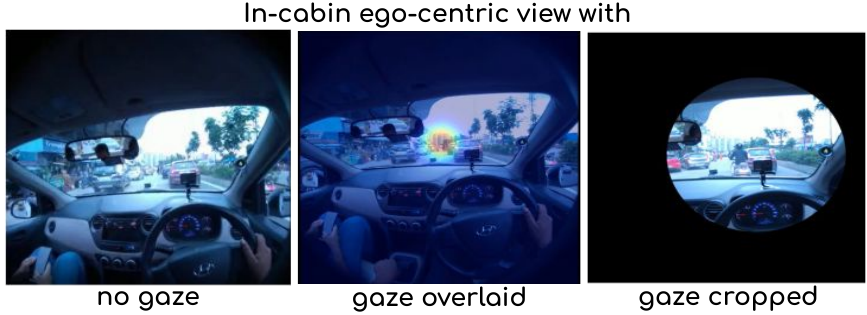}
    \caption{\textbf{Variants of gaze input.} The driver's view video is processed in the following way to show the best way to represent gaze without affecting spatial features. The gaze cropped variant ($R=350$) produces the best quantitative results.}
    \label{fig:gazecrop}
\end{figure}

\begin{figure*}[h]
    \centering
    \includegraphics[width=\linewidth]{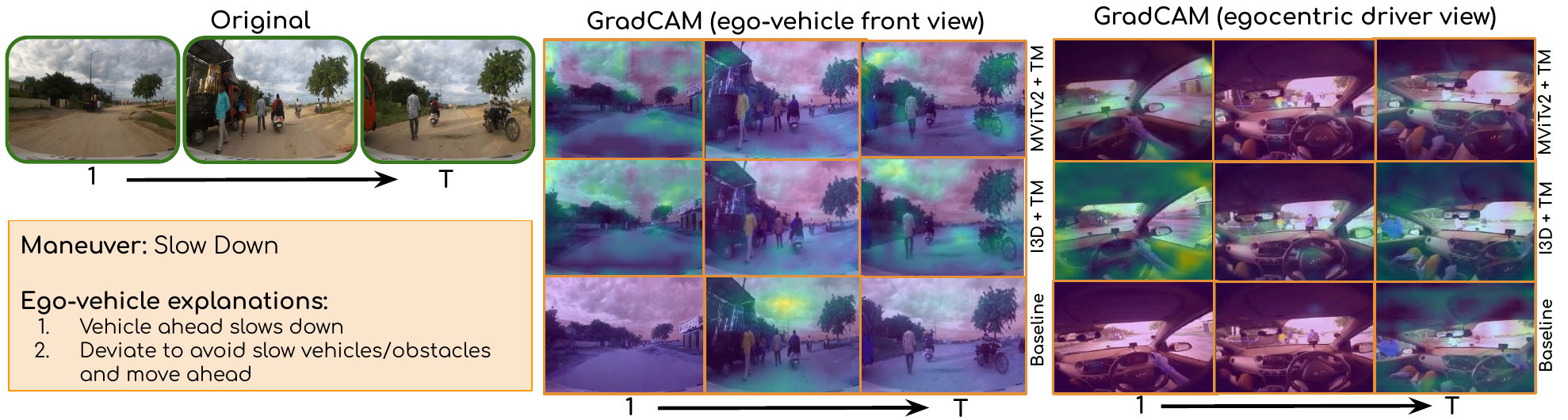}
    \caption{\textbf{GradCAM visualization on proposed method.} At $t = 1$, the activations are scattered, but as time progresses to $t = T$, the CAM gradually refines and localise on important objects. This represents how humans make decisions, which evolves over time.}
    \label{fig:gradcam}
\vspace{-.3cm}
\end{figure*}

We analyse the impact of gaze modality on explanation predictions by testing three settings: without gaze, gaze overlaid, and with gaze cropped regions, as illustrated in Fig.~\ref{fig:gazecrop}. In Table~\ref{tab:gazecrop}, we show the accuracy of identifying the gaze variant. Initially, without gaze, both action and explanation predictions are lower. Further, the gaze is being overlaid onto the driver's view, but this method adds noise to the image and deteriorates finer details. To mitigate this issue, we cropped a circular region centered on the gaze ground truth from the driver's view, experiments with various diameters (in pixels), in Table~\ref{tab:gazecrop} show improved performance for explanation by achieving an optimal score at $R=350$. However, excessively increasing the crop size reduced the concentration of relevant gaze information, reducing the explanation performance.

\subsubsection{Tradeoff between Explanation and Action Classification}

We train explanation classification jointly with action prediction to align with human reasoning. Table \ref{tab:lambda} shows that adding the auxiliary explanation loss through the scaling parameter $\lambda$ in Equation \ref{eq:loss}  boosts both explanation and action accuracy. However, excessive weight can slightly affect the action classification performance.

\begin{table}[ht]
\vspace{-1pt}
\centering
\scriptsize
\setlength{\tabcolsep}{5pt} 
\caption{\textbf{Tradeoff of Explanation Classification.} Increasing the emphasis on explanation classification leads to a decline in the action prediction performance.}
\vspace{-1pt}
\renewcommand{\arraystretch}{1.2}
\begin{tabular}{c |cc | cc | cc}
\hline
  \multirow{2}{*}{\textbf{\makecell{scaling factor ($\lambda$)}}}  & \multicolumn{2}{c|}{\textbf{Action}} & \multicolumn{4}{c}{\textbf{ego-vehicle eXplanation}} \\
\cline{2-7}
& \multicolumn{1}{c}{Acc} & \multicolumn{1}{c|}{$F_1$} & \multicolumn{1}{c}{Acc} & \multicolumn{1}{c|}{$F_1$} & \multicolumn{1}{c}{$F_1(mac)$} & \multicolumn{1}{c}{$F_1(mic)$} \\
\hline
  0 & 72.14 & 71.82 & 0 & 14.56 & 8 & 8.65 \\
  \hline
  0.01 & 71.07 & 69.94 & 0.71 & 9.87 & 3.3 & 12.95 \\
\hline
 0.1 & \textbf{74.28} & \textbf{73.43} & 5.71 & 17.36 & 10.09 & 14.48 \\
 \hline
  0.5  & 73.21 & 72.2 & 28.31 & 39.43 & \textbf{24.1} & 44.06 \\ 
 \hline
  1  & 70.35 & 69.23 & \textbf{30} & \textbf{40.02} & 22.58 & \textbf{46.4} \\ 
\hline
\end{tabular}
\vspace{-.1cm}
\label{tab:lambda}
\end{table}

\subsubsection{Effect of Temporal Cues}
\begin{figure}[h]
    \centering
    \includegraphics[width=\linewidth]{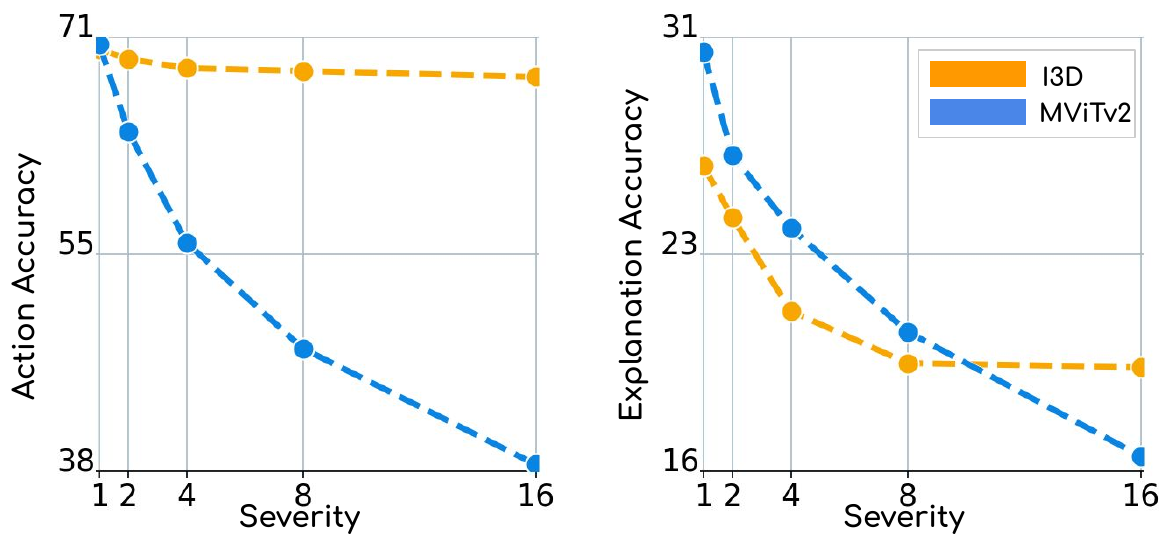}
    \caption{\textbf{Effect of temporal cues.} As the severity of frame reshuffling increases, the action and explanation accuracy of MViTv$2$ drops significantly compared to I3D. Notably, explanation accuracy drops more than action accuracy, indicating the importance of temporal cues for producing meaningful explanations.}
    \label{fig:shuffle}
\vspace{-.1cm}
\end{figure}

In  Fig.~\ref{fig:shuffle}, we show the impact of temporal cues on action and explanation performance across CNN and transformer models. Interestingly, transformers exhibit lower action prediction accuracy than CNNs, likely due to two factors: (1) CNNs rely more on spatial features and process limited temporal context, suggesting that DIP tasks can be addressed at the frame level but at the cost of explainability, and (2) transformers undergo stronger regularisation to prevent learning spurious correlations from noisy data, with the help of random shuffling as discussed in Sec.~\ref{sec:method}. This random shuffling disrupts temporal order, affecting transformer performance, while CNNs remain unaffected since they are less reliant on temporal information. The severity $s$ in Fig.~\ref{fig:shuffle} represents the degree of reshuffling. 

Let \( T \) be the total number of frames in the video. We first divide it into $16$ equal segments, each containing $\ell = \frac{T}{16}$ frames. Based on the severity parameter \( s \), we merge every \( s \) consecutive segments, forming $M = \frac{16}{s}
$ merged segments. From each merged segment, we uniformly sample \( s \) frames:

\[
F_{i} = \left\{ f_{i,1}, f_{i,2}, \dots, f_{i,s} \right\}, \quad i = 1, \dots, M
\]

where each sampled frame \( f_{i,j} \) is selected uniformly from the \( i \)$^{th}$ merged segment. The total number of sampled frames sums up to:

\[
\sum_{i=1}^{M} |F_{i}| = 16.
\]
This indicates that our explanation annotations are influenced by temporal dependencies.

\begin{table}
\centering
\caption{\textbf{Gaze modality input variants.} Having the gaze cropped regions is better than the usual way of overlaying the gaze in the DIP task.}

\label{tab:gazecrop}
\footnotesize 
\setlength{\tabcolsep}{5pt} 

\renewcommand{\arraystretch}{1.2}
\begin{tabular}{c |cc | cc | cc}
\hline

  \multirow{2}{*}{\makecell{\textbf{Variants}}}  & \multicolumn{2}{c|}{\textbf{Action}} & \multicolumn{4}{c}{\textbf{ego-vehicle eXplanation}} \\
\cline{2-7}
& \multicolumn{1}{c}{Acc} & \multicolumn{1}{c|}{$F_1$} & \multicolumn{1}{c}{Acc} & \multicolumn{1}{c|}{$F_1$} & \multicolumn{1}{c}{$F_1(mac)$} & \multicolumn{1}{c}{$F_1(mic)$} \\
\hline
 no gaze & 68.11 & 67.94 & 8.77 & 17.52 & 9.37 & 22.51 \\
overlaid & 71.57 & 70.57 & 14.03 & 22.6 & 11.86 & 28.55 \\
 50 & 67.85 & 67.51 & 14.28 & 24.24 & 11.34 & 28.8 \\
 150 & 70.21 & 69.2 & 20.63 & 28.46 & 15.22 & 34.18 \\
250 & 72.63 & 72.42  & 23.74 & 33.36 & 17.59 & 40.13 \\
 350 & 74.09 & 73.47 & \textbf{26.42} & \textbf{36.73} & 18.53 & \textbf{43.49} \\
 450 & \textbf{74.64} & 73.74 & 25.26 & 36.01 & \textbf{19.18} & 43.46 \\
 550 & 74.28 & \textbf{73.75} & 24.64 & 35.67 & 17.39 & 43.32 \\
\hline
\end{tabular}
\end{table}

\subsection{Qualitative Analysis}

\subsubsection{GradCAM Visualization}
As shown in Fig.~\ref{fig:gradcam}, the vehicle's current maneuver is predicted as ``Slow Down", with intermediate outputs providing ego-vehicle explanations. This demonstrates that LTM and LCBM effectively focus activations on relevant features that align with the predicted explanations. In contrast, the baseline CBM, without late averaging and LTM, results in more dispersed, global activations, making it less interpretable. 

\subsubsection{Label-anchored Multi-label t-SNE Visualization}
Visualizing explanations in the feature space is essential for understanding what the DIP model has learned in maneuver prediction. When dealing with multi-label explanations, techniques like t-SNE help interpret the latent space and reveal relationships between different explanations. However, since t-SNE is not directly suited for multi-label classification, we introduce explanations as anchor points within the latent space. This approach offers two key benefits: (1) anchor points highlight the degree of correlation among different explanations, and (2) individual video features are positioned in alignment with and in close proximity to their relevant anchor points, ensuring a more interpretable representation of the learned features.

To formalize, let $\mathbf{z'}_i \in \mathbb{R}^d$ be the backbone feature vector for the $i$-th sample, for $i = 1, \ldots, T$. For each explanation $k$, define a mask indicator $s_i^{(k)}$ as follows,
\[
s_i^{(k)} =
\begin{cases}
1, & \text{if class } k \text{ is activated for sample } i, \\
0, & \text{otherwise}.
\end{cases}
\]
The aggregated feature representation for the explanation $k$, denoted by $\bar{\mathbf{z}}_k$, is calculated by applying the mask to the features and then averaging over the samples where the mask is active,
\[
\bar{\mathbf{z}}_k = \frac{\sum_{i=1}^{T} s_i^{(k)} \mathbf{z'}_i}{\sum_{i=1}^{T} s_i^{(k)}}, 
\]

This aggregated feature $\bar{\mathbf{z}}_k$ serves as an anchor in the 2D t-SNE space for explanation $k$. This makes the feature space more interpretable, revealing degree of causal relationships learned between explanations.

\begin{figure}[h]
    \centering
    \includegraphics[width=\linewidth]{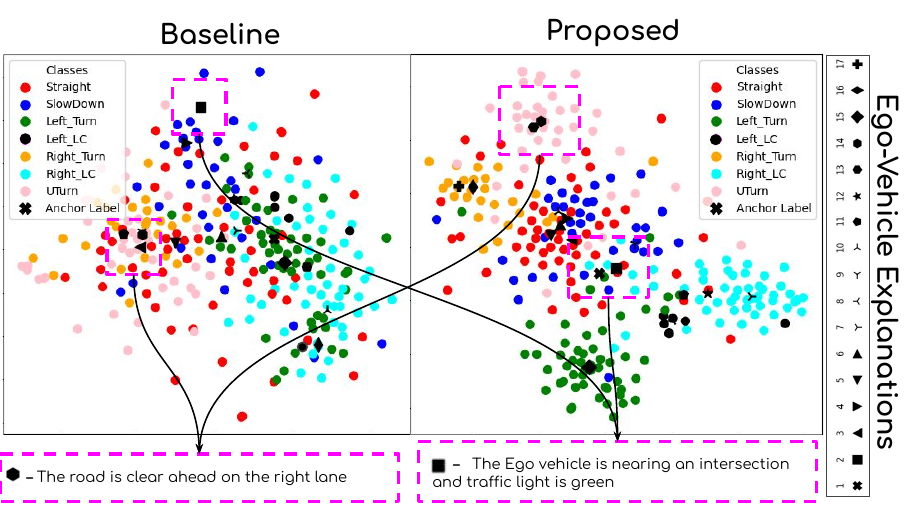}
    
    \caption{\textbf{Label-anchored multi-label t-SNE.} Colored dots represent clusters of individual video features. \textbf{Left}:  Baseline model exhibiting a poorly disentangled representation space. \textbf{Right}: Proposed method demonstrating improved separation of explanation symbols, with a stronger causal correlation to the videos. The square marker is positioned at the center, representing a feature commonly observed across all videos. The hexagon indicates an explanation learned in scenarios where a U-turn is performed, as a right turn and a right lane change always accompany it.}
    \label{fig:msne}
\end{figure}

Fig.~\ref{fig:msne} illustrates the $17$ explanation anchors, each marked with distinct shapes. This visualization highlights that semantically related explanations tend to form clusters, while individual video features (depicted as colored points) are positioned near their corresponding explanation anchors. For example, if a video contains both ``traffic light is green" and ``left turn coming ahead," its feature representation will be located near both respective anchors, reflecting the model's learned associations. 

\noindent\textbf{Limitations and future scope:} Our proposed method generates high-level explanations while preserving faithful feature attributes. However, GradCAM activations are predominantly observed in the forward direction. This occurs because we assume that important objects are only considered if they are visible from both views, i.e, the driver's gaze should guide, where the model should focus on from the front view. Consequently, token merging relies on the assumption that the video frames are at least partially aligned. An interesting direction for future work would be to investigate the effects of explicitly aligning both views before performing token merging using techniques like homography.

\section{Conclusion}
\label{sec:conclusion}
This work introduces a novel paradigm for conceptual interpretability in driving maneuver prediction. We developed a comprehensive multi-modal dataset incorporating human-understandable explanations to help create interpretable models within the autonomous driving systems. Our analysis of existing architectures revealed that transformer models excel at generating explanations due to their inherent temporal bias. Leveraging this insight, we proposed VCBM, a model that merges spatio-temporal features to predict localised explanations and reliably represents them through post hoc feature attribution methods. Additionally, our feature-level visualization approach effectively elucidates the causal correlations among explanations, enhancing driver intention prediction systems' overall transparency and reliability.

\hfill \break
\noindent
\textbf{Acknowledgment.} The project was supported by iHub-Data and Mobility at IIIT Hyderabad.

{
    \small
    \bibliographystyle{ieeenat_fullname}
    \bibliography{main}
}

\end{document}